\documentclass[10pt,twocolumn,letterpaper]{article}
\usepackage{cvpr} 
\usepackage[accsupp]{axessibility}  

\usepackage{graphicx}
\usepackage{amsmath}
\usepackage{amssymb}
\usepackage{booktabs}
\usepackage{multirow}
\usepackage{enumitem}
\usepackage{array}
\usepackage{makecell}
\usepackage{microtype}
\usepackage{caption}
\usepackage{subcaption}
\usepackage{circledsteps}
\usepackage[numbers,sort,compress]{natbib}
\usepackage[pagebackref,breaklinks,colorlinks,bookmarks=false]{hyperref}

\usepackage{optidef}

\usepackage[capitalize]{cleveref}
\crefname{section}{Sec.}{Secs.}
\Crefname{section}{Section}{Sections}
\Crefname{table}{Table}{Tables}
\crefname{table}{Tab.}{Tabs.}
\usepackage{booktabs,caption}
\def\cvprPaperID{6224} 
\def\confName{CVPR}
\def\confYear{2023}

\begin{document}
\def\mA{\mathcal{A}}
\def\mB{\mathcal{B}}
\def\mC{\mathcal{C}}
\def\mD{\mathcal{D}}
\def\mE{\mathcal{E}}
\def\mF{\mathcal{F}}
\def\mG{\mathcal{G}}
\def\mH{\mathcal{H}}
\def\mI{\mathcal{I}}
\def\mJ{\mathcal{J}}
\def\mK{\mathcal{K}}
\def\mL{\mathcal{L}}
\def\mM{\mathcal{M}}
\def\mN{\mathcal{N}}
\def\mO{\mathcal{O}}
\def\mP{\mathcal{P}}
\def\mQ{\mathcal{Q}}
\def\mR{\mathcal{R}}
\def\mS{\mathcal{S}}
\def\mT{\mathcal{T}}
\def\mU{\mathcal{U}}
\def\mV{\mathcal{V}}
\def\mW{\mathcal{W}}
\def\mX{\mathcal{X}}
\def\mY{\mathcal{Y}}
\def\mZ{\mathcal{Z}} 

\def\bbN{\mathbb{N}} 
\def\bbR{\mathbb{R}} 
\def\bbP{\mathbb{P}} 
\def\bbQ{\mathbb{Q}} 
\def\bbE{\mathbb{E}}

\def\1n{\mathbf{1}_n}
\def\0{\mathbf{0}}
\def\1{\mathbf{1}}

\def\A{{\bf A}}
\def\B{{\bf B}}
\def\C{{\bf C}}
\def\D{{\bf D}}
\def\E{{\bf E}}
\def\F{{\bf F}}
\def\G{{\bf G}}
\def\H{{\bf H}}
\def\I{{\bf I}}
\def\J{{\bf J}}
\def\K{{\bf K}}
\def\L{{\bf L}}
\def\M{{\bf M}}
\def\N{{\bf N}}
\def\O{{\bf O}}
\def\P{{\bf P}}
\def\Q{{\bf Q}}
\def\R{{\bf R}}
\def\S{{\bf S}}
\def\T{{\bf T}}
\def\U{{\bf U}}
\def\V{{\bf V}}
\def\W{{\bf W}}
\def\X{{\bf X}}
\def\Y{{\bf Y}}
\def\Z{{\bf Z}}

\def\a{{\bf a}}
\def\b{{\bf b}}
\def\c{{\bf c}}
\def\d{{\bf d}}
\def\e{{\bf e}}
\def\f{{\bf f}}
\def\g{{\bf g}}
\def\h{{\bf h}}
\def\i{{\bf i}}
\def\j{{\bf j}}
\def\k{{\bf k}}
\def\l{{\bf l}}
\def\m{{\bf m}}
\def\n{{\bf n}}
\def\o{{\bf o}}
\def\p{{\bf p}}
\def\q{{\bf q}}
\def\r{{\bf r}}
\def\s{{\bf s}}
\def\t{{\bf t}}
\def\u{{\bf u}}
\def\v{{\bf v}}
\def\w{{\bf w}}
\def\x{{\bf x}}
\def\y{{\bf y}}
\def\z{{\bf z}}

\def\balpha{\mbox{\boldmath{$\alpha$}}}
\def\bbeta{\mbox{\boldmath{$\beta$}}}
\def\bdelta{\mbox{\boldmath{$\delta$}}}
\def\bgamma{\mbox{\boldmath{$\gamma$}}}
\def\blambda{\mbox{\boldmath{$\lambda$}}}
\def\bsigma{\mbox{\boldmath{$\sigma$}}}
\def\btheta{\mbox{\boldmath{$\theta$}}}
\def\bomega{\mbox{\boldmath{$\omega$}}}
\def\bxi{\mbox{\boldmath{$\xi$}}}
\def\bnu{\mbox{\boldmath{$\nu$}}}                                  
\def\bphi{\mbox{\boldmath{$\phi$}}}
\def\bmu{\mbox{\boldmath{$\mu$}}}

\def\bDelta{\mbox{\boldmath{$\Delta$}}}
\def\bOmega{\mbox{\boldmath{$\Omega$}}}
\def\bPhi{\mbox{\boldmath{$\Phi$}}}
\def\bLambda{\mbox{\boldmath{$\Lambda$}}}
\def\bSigma{\mbox{\boldmath{$\Sigma$}}}
\def\bGamma{\mbox{\boldmath{$\Gamma$}}}
                                  
\newcommand{\myprob}[1]{\mathop{\mathbb{P}}_{#1}}

\newcommand{\myexp}[1]{\mathop{\mathbb{E}}_{#1}}

\newcommand{\mydelta}[1]{1_{#1}}

\newcommand{\myminimum}[1]{\mathop{\textrm{minimum}}_{#1}}
\newcommand{\mymaximum}[1]{\mathop{\textrm{maximum}}_{#1}}    
\newcommand{\mymin}[1]{\mathop{\textrm{minimize}}_{#1}}
\newcommand{\mymax}[1]{\mathop{\textrm{maximize}}_{#1}}
\newcommand{\mymins}[1]{\mathop{\textrm{min.}}_{#1}}
\newcommand{\mymaxs}[1]{\mathop{\textrm{max.}}_{#1}}  
\newcommand{\myargmin}[1]{\mathop{\textrm{argmin}}_{#1}} 
\newcommand{\myargmax}[1]{\mathop{\textrm{argmax}}_{#1}} 
\newcommand{\myst}{\textrm{s.t. }}

\newcommand{\denselist}{\itemsep -1pt}
\newcommand{\sparselist}{\itemsep 1pt}

\definecolor{pink}{rgb}{0.9,0.5,0.5}
\definecolor{purple}{rgb}{0.5, 0.4, 0.8}   
\definecolor{gray}{rgb}{0.3, 0.3, 0.3}
\definecolor{mygreen}{rgb}{0.2, 0.6, 0.2}

\newcommand{\cyan}[1]{\textcolor{cyan}{#1}}
\newcommand{\red}[1]{\textcolor{red}{#1}}  
\newcommand{\blue}[1]{\textcolor{blue}{#1}}
\newcommand{\magenta}[1]{\textcolor{magenta}{#1}}
\newcommand{\pink}[1]{\textcolor{pink}{#1}}
\newcommand{\green}[1]{\textcolor{green}{#1}} 
\newcommand{\gray}[1]{\textcolor{gray}{#1}}    
\newcommand{\mygreen}[1]{\textcolor{mygreen}{#1}}    
\newcommand{\purple}[1]{\textcolor{purple}{#1}}       

\definecolor{greena}{rgb}{0.4, 0.5, 0.1}
\newcommand{\greena}[1]{\textcolor{greena}{#1}}

\definecolor{bluea}{rgb}{0, 0.4, 0.6}
\newcommand{\bluea}[1]{\textcolor{bluea}{#1}}
\definecolor{reda}{rgb}{0.6, 0.2, 0.1}
\newcommand{\reda}[1]{\textcolor{reda}{#1}}

\def\changemargin#1#2{\list{}{\rightmargin#2\leftmargin#1}\item[]}
\let\endchangemargin=\endlist
                                               
\newcommand{\cm}[1]{}

\newcommand{\mhoai}[1]{{\color{magenta}\textbf{[MH: #1]}}}

\newcommand{\mtodo}[1]{{\color{red}$\blacksquare$\textbf{[TODO: #1]}}}
\newcommand{\myheading}[1]{\vspace{1ex}\noindent \textbf{#1}}
\newcommand{\htimesw}[2]{\mbox{$#1$$\times$$#2$}}


\newif\ifshowsolution
\showsolutiontrue

\ifshowsolution  
\newcommand{\Comment}[1]{\paragraph{\bf $\bigstar $ COMMENT:} {\sf #1} \bigskip}
\newcommand{\Solution}[2]{\paragraph{\bf $\bigstar $ SOLUTION:} {\sf #2} }
\newcommand{\Mistake}[2]{\paragraph{\bf $\blacksquare$ COMMON MISTAKE #1:} {\sf #2} \bigskip}
\else
\newcommand{\Solution}[2]{\vspace{#1}}
\fi

\newcommand{\truefalse}{
\begin{enumerate}
	\item True
	\item False
\end{enumerate}
}

\newcommand{\yesno}{
\begin{enumerate}
	\item Yes
	\item No
\end{enumerate}
}

\newcommand{\Sref}[1]{Sec.~\ref{#1}}
\newcommand{\Eref}[1]{Eq.~(\ref{#1})}
\newcommand{\Fref}[1]{Fig.~\ref{#1}}
\newcommand{\Tref}[1]{Table~\ref{#1}}

\definecolor{mydarkblue}{rgb}{0,0.08,1}
\definecolor{mydarkgreen}{rgb}{0.02,0.6,0.02}
\definecolor{myred}{rgb}{1.0,0.0,0.0}
\newcommand{\khoi}[1]{\textcolor{mydarkblue}{[Khoi: #1]}}
\newcommand{\tuan}[1]{\textcolor{mydarkgreen}{[Tuan: #1]}}
\newcommand{\son}[1]{\textcolor{myred}{[Son: #1]}}

\title{Universal Representations for Classification-enhanced Lossy Compression}

\author{Nam Nguyen\\Oregon State University, Oregon, United States\\
{\tt\small \{nguynam4\}@oregonstate.edu}
}

\maketitle
\thispagestyle{plain}
\pagestyle{plain}
\begin{abstract}
In lossy compression, the classical tradeoff between compression rate and reconstruction distortion has traditionally guided algorithm design. However, Blau and Michaeli \cite{blau2019rethinking} introduced a generalized framework, known as the rate-distortion-perception (RDP) function, incorporating perceptual quality as an additional dimension of evaluation. More recently, the rate-distortion-classification (RDC) function was investigated in \cite{Wang2024}, evaluating compression performance by considering classification accuracy alongside distortion. In this paper, we explore universal representations, where a single encoder is developed to achieve multiple decoding objectives across various distortion and classification (or perception) constraints. This universality avoids retraining encoders for each specific operating point within these tradeoffs. Our experimental validation on the MNIST dataset indicates that a universal encoder incurs only minimal performance degradation compared to individually optimized encoders for perceptual image compression tasks, aligning with prior results from \cite{UniversalRDPs}. Nonetheless, we also identify that in the RDC setting, reusing an encoder optimized for one specific classification-distortion tradeoff leads to a significant distortion penalty when applied to alternative points.
\end{abstract}

\vspace{-12pt}
\section{Introduction}
\label{sec:intro}

\begin{figure*}[h]
\centering
\vspace{-0.2cm}
\includegraphics[width=0.6\textwidth]{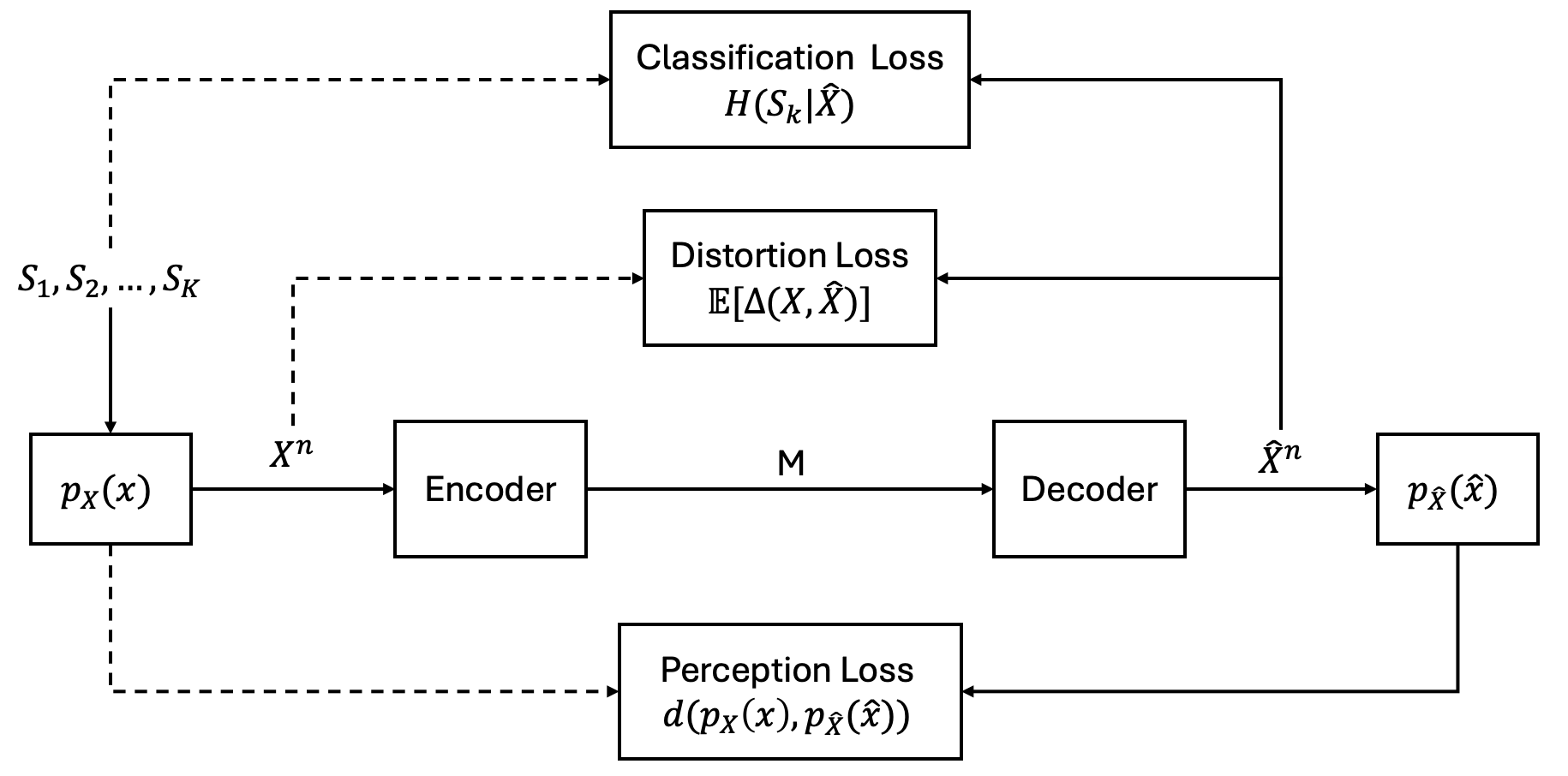}
\caption{Schematic representation of a task-oriented lossy compression framework.}
\label{Lossy_Compression_Framework}
\end{figure*}
Lossy compression involves reconstructing data from compressed representations with acceptable levels of distortion, typically measured using metrics such as mean squared error (MSE), PSNR, and SSIM \cite{wang2003multiscale,wang2004image}. While classical rate-distortion theory seeks to minimize distortion for a given compression rate, recent studies have demonstrated that lower distortion does not necessarily imply higher perceptual quality \cite{blau2018perception, agustsson2019generative}. Addressing this discrepancy, Blau and Michaeli introduced the rate-distortion-perception (RDP) framework \cite{blau2019rethinking}, which incorporates perceptual quality—measured through the divergence between source and reconstruction distributions—as an additional dimension of evaluation. The emergence of generative adversarial networks (GANs) \cite{goodfellow2014generative} has facilitated this perceptual enhancement even at low bitrates \cite{tschannen2018deep}, highlighting a clear tradeoff between perceptual quality and distortion.

Additionally, recent work has explored the integration of classification performance into compression objectives, resulting in the rate-distortion-classification (RDC) tradeoff \cite{Wang2024,Zhang2023}. Here, improved classification performance is often achieved at the cost of increased distortion or reduced perceptual quality.

In this paper, we investigate the concept of \textit{universal representations}, where a single encoder is used to support multiple decoding objectives across various distortion-classification (or distortion-perception) tradeoff points. We demonstrate through experiments on the MNIST dataset that universal encoders achieve performance close to specialized encoders in perceptual image compression tasks, confirming prior findings \cite{UniversalRDPs}. However, we also highlight that reusing encoders trained for a particular classification-distortion tradeoff introduces a notable distortion penalty when applied to other points. Our findings suggest practical ways to simplify and streamline the training of deep-learning-based compression systems by reducing the need for specialized encoders. Throughout this study, we focus on scenarios where the compression rate is fixed.

\section{Related Works}
Image quality assessment typically involves full-reference metrics such as MSE and SSIM \cite{wang2004image}, or no-reference metrics like BRISQUE and Fréchet Inception Distance \cite{heusel2017gans}. Broadly, full-reference metrics quantify distortion, whereas no-reference metrics measure perceptual quality. Recently, GAN-based methods have enhanced perceptual realism by leveraging discriminators trained to estimate statistical divergences \cite{arjovsky2017wasserstein}.

Rate-distortion theory provides the foundational framework for analyzing lossy compression \cite{cover1999elements}, influencing both representation learning and generative modeling \cite{tschannen2018recent}. Distribution-preserving lossy compression has also been studied within classical information theory \cite{zamir2001natural}. Recent approaches have integrated GAN regularization into compressive autoencoders to reduce artifacts, significantly improving perceptual quality even at low bitrates \cite{mentzer2020high}.

Incorporating classification performance into compression objectives has resulted in notable tradeoffs between distortion, perception, and classification accuracy \cite{CDP,Wang2024}. Such studies highlight that improved classification accuracy frequently leads to increased distortion or lower perceptual quality.

\section{Problem Formulation}
\label{sec:problem_formulation}
Consider a source generating observable data $X \sim p_X(x)$ associated with several underlying but unobserved labels $S_1,\dots,S_K$. These labels and the observation $X$ follow a joint distribution $p_{X,S}(x,s_1,\dots,s_K)$. For instance, $X$ could represent speech data, with labels such as spoken content, speaker identity, or speaker gender.

As depicted in Fig. \ref{Lossy_Compression_Framework}, a lossy compression scheme consists of an encoder-decoder pair. Given an independent and identically distributed (i.i.d.) sequence $X_1, X_2, \dots, X_n \sim p_X(x)$: The encoder maps the source sequence $X^n$ into a compressed representation $M \in \{1, 2, \dots, 2^{nR}\}$ at rate $R$ bits. The decoder reconstructs data $\hat{X}^n$ from the compressed message $M$.

\myheading{Distortion constraint.} The reconstruction must satisfy a distortion limit:
\begin{equation}
    \mathbb{E}[\Delta(X, \hat{X})] \leq D,
\end{equation}
where $\Delta(\cdot,\cdot)$ measures distortion, e.g., MSE or Hamming distance \cite{cover1999elements}, and the expectation is taken over the joint distribution of $X$ and $\hat{X}$.

\myheading{Classification constraint.} The reconstructed data should preserve sufficient information for accurate inference of labels:
\begin{equation}
    H(S_k|\hat{X}) \leq C_k, \quad \forall k \in [K],
\end{equation}
where $H(S_k|\hat{X})$ denotes conditional entropy, limiting uncertainty about each label given the reconstructed data \cite{Wang2024}.

\myheading{Perception constraint.} Perceptual quality measures how natural and realistic reconstructed images appear to human observers, distinguishing them from artificially generated or processed images \cite{moorthy2011blind_perceptual}. To formally quantify perceptual quality, we impose a constraint based on a divergence measure between the distributions of original and reconstructed data, following established practices \cite{PD-tradeoff,RethinkingRDP}:
\begin{equation}
    d(p_X, p_{\hat{X}}) \leq P,
\end{equation}
where $d(\cdot,\cdot)$ denotes a statistical divergence measure such as total-variation (TV) divergence or Kullback-Leibler (KL) divergence.

To characterize achievable rates under these constraints, we define the information rate-distortion-perception-classification (RDPC) function \cite{CDP}:
\begin{mini!}|s|[2]
{p_{\hat{X}|X}}
{I(X;\hat{X})}
{\label{RDPC}}
{R(D,P,\mathbf{C}) =}
\addConstraint{\mathbb{E}[\Delta(X, \hat{X})]}{\leq D}
\addConstraint{d(p_X,p_{\hat{X}})}{\leq P}
\addConstraint{H(S_k|\hat{X})}{\leq C_k, \quad \forall k\in[K],}
\end{mini!}
where $\mathbf{C}=(C_1,\dots,C_K)$ represents the constraints on classification uncertainty.

\section{Rate-Distortion-Perception/Classification Representations}
\label{sec:rate_distortion_classification}
In lossy compression, an encoder-decoder pair optimizes the tradeoffs among compression rate, distortion, and additional constraints like perceptual quality or classification accuracy. We formally define two essential functions for these tradeoffs: the information rate-distortion-perception and information rate-distortion-classification.

\myheading{Information rate-distortion-perception function.} Incorporating perceptual quality into the classical rate-distortion framework, the RDP function is defined as:
\begin{mini!}|s|[2]
{p_{\hat{X}|X}}
{I(X;\hat{X})}
{\label{RDP}}
{R(D,P) =}
\addConstraint{\mathbb{E}[\Delta(X, \hat{X})]}{\leq D}
\addConstraint{d(p_X,p_{\hat{X}})}{\leq P.}
\end{mini!}
where $\Delta(\cdot,\cdot)$ measures reconstruction distortion, and $d(p_X,p_{\hat{X}})$ is the divergence between original and reconstructed distributions. The RDP function is monotonically non-increasing and convex \cite{blau2019rethinking}.

\myheading{Information rate-distortion-classification function.} For tasks emphasizing classification accuracy, the RDC function characterizes the optimal tradeoff:
\begin{mini!}|s|[2]
{p_{\hat{X}|X}}
{I(X;\hat{X})}
{\label{RDC}}
{R(D,C) =}
\addConstraint{\mathbb{E}[\Delta(X, \hat{X})]}{\leq D}
\addConstraint{H(S|\hat{X})}{\leq C,}
\end{mini!}
where $H(S|\hat{X})$ limits uncertainty about labels. Similar to RDP, the RDC function is monotonically non-increasing and convex \cite{Wang2024}.

\section{Universal Representations for RDC}
\label{sec:universal_representatons}
\begin{figure}[!htbp]
\centering
\includegraphics[width=0.42\textwidth]{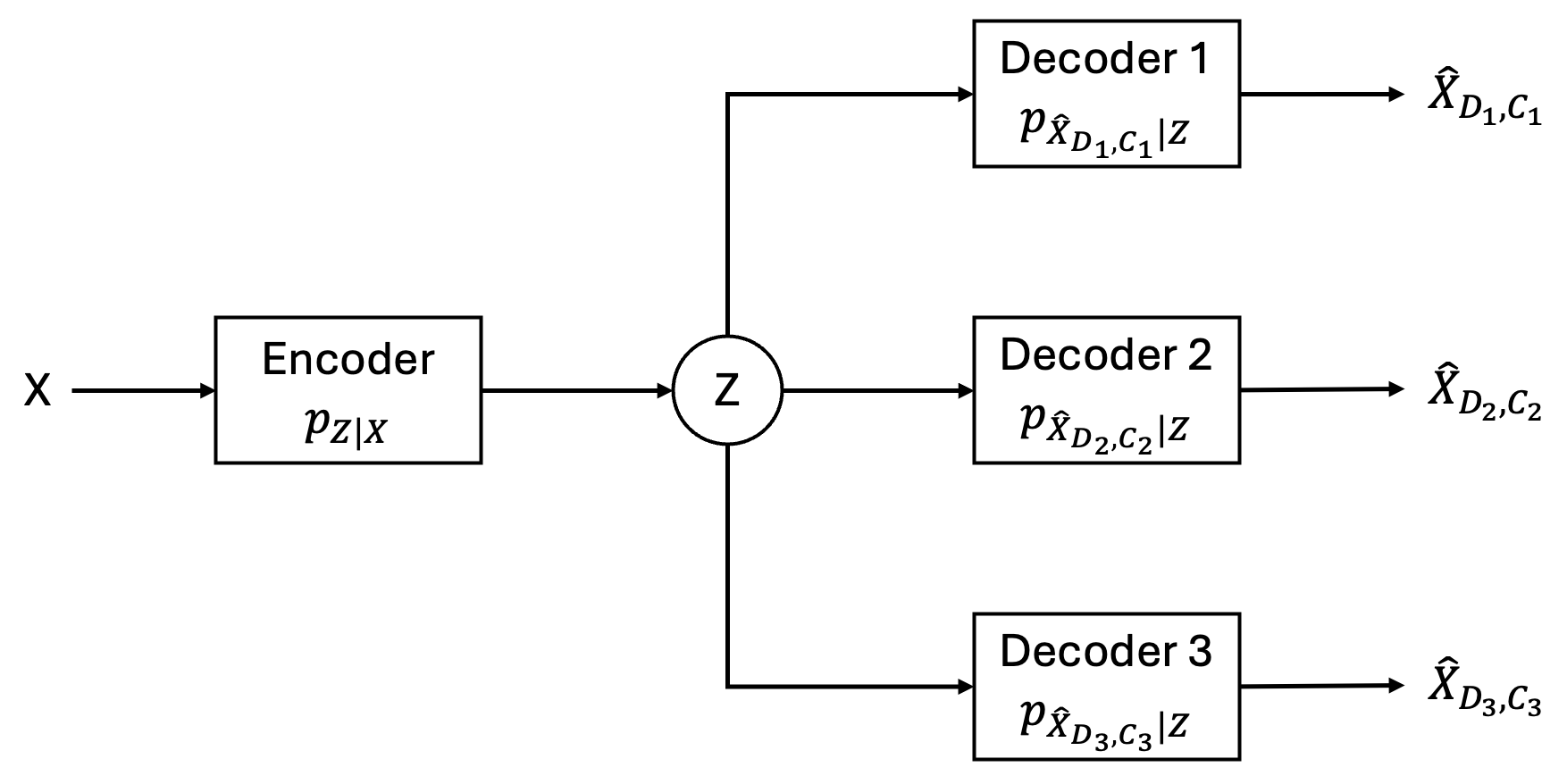}
\caption{Illustration of the universal representation framework.}
\label{fig:Universal}
\end{figure}
The concept of universal representations, initially introduced in \cite{UniversalRDPs} for the rate-distortion-perception framework, aims to design a single encoder that supports multiple decoding objectives. Here, we extend this concept to the rate-distortion-classification scenario.

Consider a fixed encoder that generates a universal representation, from which various decoders can reconstruct data satisfying multiple distortion-classification constraints $(D,C)$ as shown in Figure \ref{fig:Universal}. Formally, the information universal rate-distortion-classification function is defined as:
\begin{equation}
R(\Theta) = \inf_{p_{Z|X}\in\mathcal{P}_{Z|X}(\Theta)} I(X;Z),
\end{equation}
where $\mathcal{P}_{Z|X}(\Theta)$ includes all encoding schemes allowing decoders to satisfy constraints $(D,C) \in \Theta$.

The \textit{rate penalty} from using a single universal encoder instead of specialized encoders is:
\begin{equation}
A(\Theta)=R(\Theta)-\sup_{(D,C)\in\Theta}R(D,C).
\end{equation}
Ideally, if we define $\Omega(R)=\{(D,C): R(D,C)\leq R\}$, the rate penalty $A(\Omega(R))$ should be minimal (preferably zero). This implies that one encoder can efficiently achieve multiple classification-distortion constraints without incurring additional encoding cost, significantly simplifying the system design and reducing computational overhead.

\section{Experiments}
\label{sec:experiments}
\section{Universal Representation for Lossy Compression}
The rate-distortion-classification tradeoff arises naturally when integrating classification objectives into deep-learning-based image compression \cite{Wang2024}. Typically, each desired RDC setting requires training an individual encoder-decoder pair. However, retraining entire models for every specific objective is inefficient. Therefore, we consider reusing a pre-trained encoder and adjusting only the decoder—termed as \textit{universal} models. In contrast, models fully retrained for each objective are called \textit{end-to-end} models. Our goal is to quantify the distortion and classification performance penalties when reusing encoders compared to specialized training.

\begin{figure*}[!htbp]
\center \includegraphics[width=0.7\textwidth]{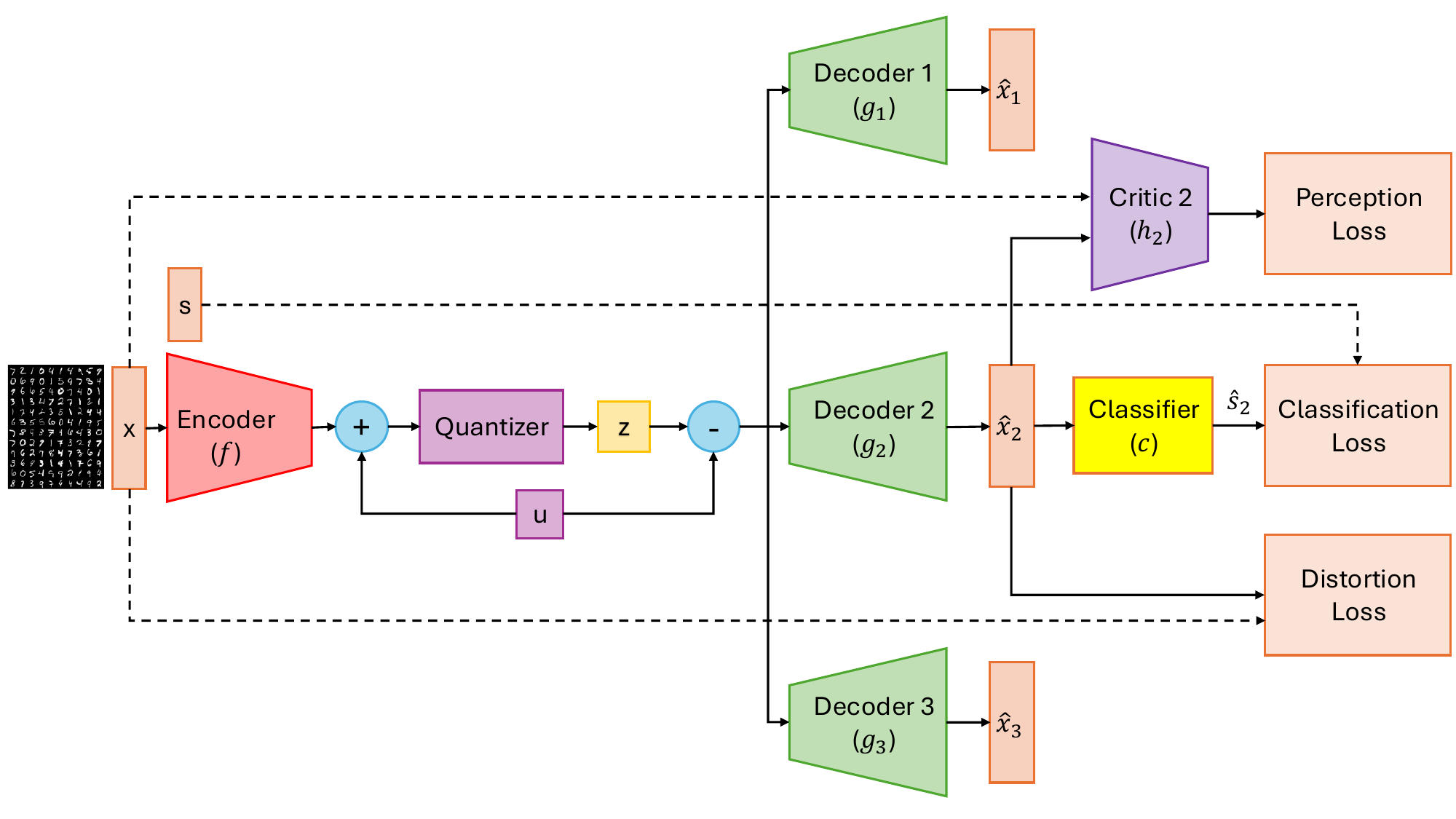}
\caption{Diagram of the experimental framework for the universal representation model. Initially, an encoder network $f$ is trained to achieve a predetermined balance between classification accuracy and reconstruction distortion (alternatively, perception and distortion). After training, the encoder's parameters are fixed. Subsequently, multiple specialized decoders $\{g_i\}$ are independently optimized, each targeting distinct trade-off criteria using the fixed representation $z$ generated by encoder $f$. A shared source of randomness $u$ is accessible to both sender and receiver to facilitate universal quantization. Additionally, dedicated critic networks $\{h_i\}$ are concurrently trained alongside each decoder to enhance perceptual quality. A pre-trained classifier network ($C$) is utilized for evaluating classification performance.}
\label{fig:Scheme}
\end{figure*}

\begin{figure*}[t] 
    \centering
    \subfloat[CDR for MNIST.]{%
        \includegraphics[width=0.37\textwidth]{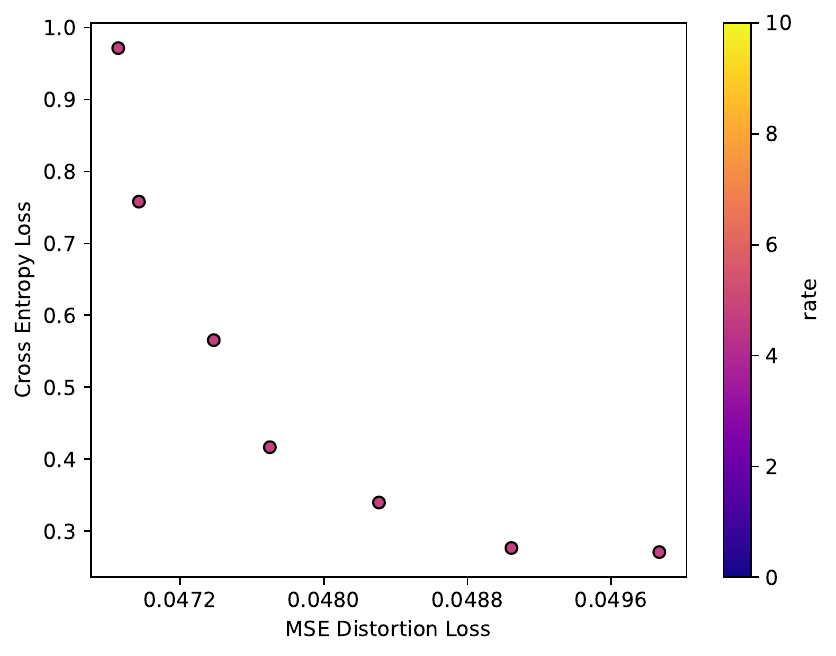}%
        \label{fig:CDR_MNIST}
    }
    \hfill
    \subfloat[RDC with multiple rates for MNIST.]{%
        \includegraphics[width=0.37\textwidth]{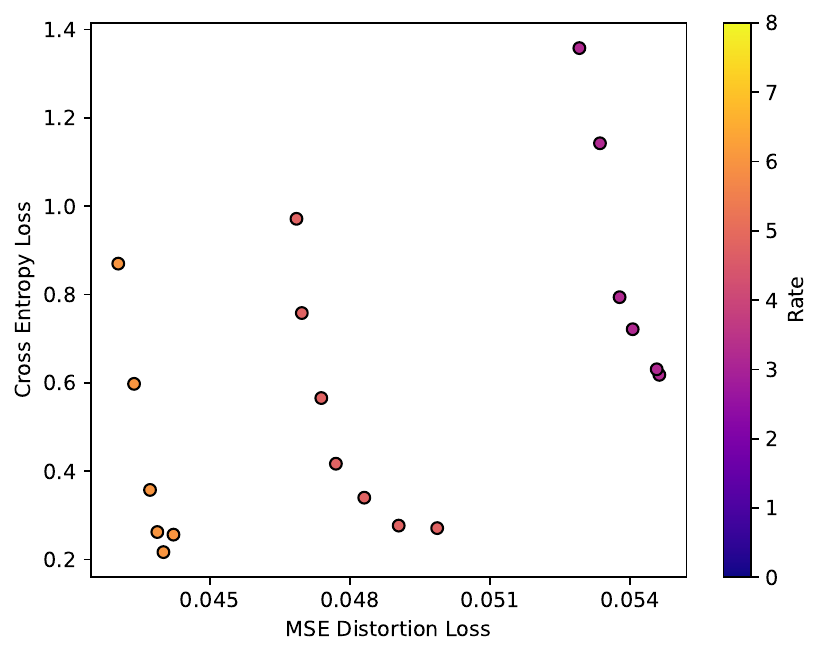}%
        \label{fig:RDC_Multiple_Rates_MNIST}
    }
    \caption{Classification-distortion-rate functions along various rates for the MNIST dataset, illustrating the tradeoff between rate, distortion, and classification.}
    \label{fig:classification_rate_distortion}
\end{figure*}

\begin{figure}[h] 
    \centering
    \includegraphics[width=0.37\textwidth]{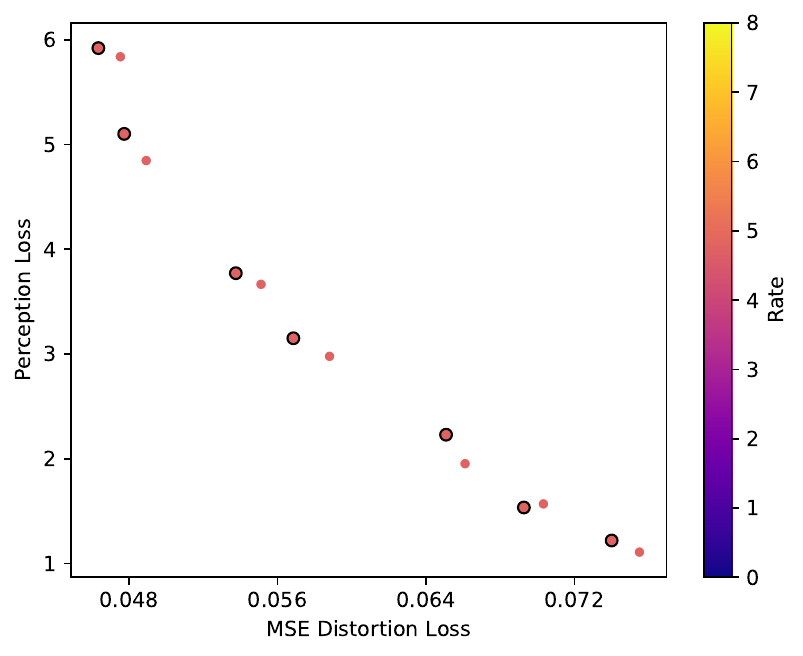}
    \caption{Perception-distortion-rate functions evaluated at a fixed rate of $R = 4.75$ on the MNIST dataset. Points highlighted with black outlines indicate results obtained from end-to-end trained encoder-decoder models tailored specifically to particular perception-distortion targets. All other points represent outcomes from universal models, in which decoders are trained separately using representations from an encoder fixed at low perceptual distortion ($\lambda_p=0.015$). The universal models closely match the performance of the jointly trained models across the entire range of trade-offs.}
    \label{fig:PDR_Comparision_MNIST_1}
\end{figure}

\begin{figure}[h] 
    \centering
    \includegraphics[width=0.37\textwidth]{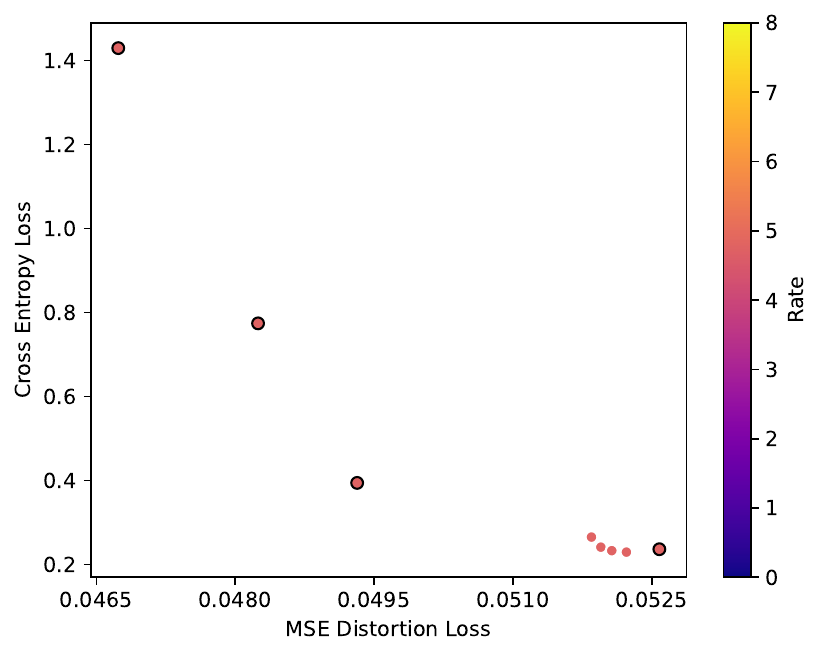}
    \caption{Classification-distortion-rate curves at a fixed rate $R = 4.75$ for the MNIST dataset. Points with black outlines correspond to results from encoder-decoder models trained jointly end-to-end for specific classification-distortion objectives. All other points depict universal model outcomes, where decoders are optimized using representations from an encoder fixed at low classification loss ($\lambda_c=0.015$). The universal models show cross-entropy losses nearly identical to the frozen encoder baseline; however, a notable distortion gap remains between universal and end-to-end models. It is suggested to utilize $\lambda^1_c$ at a different scaling factor from $\lambda_c$ for better alignment.}
    \label{fig:CDR_Comparision_MNIST_1}
\end{figure}

\begin{figure}[h] 
    \centering
    \includegraphics[width=0.45\textwidth]{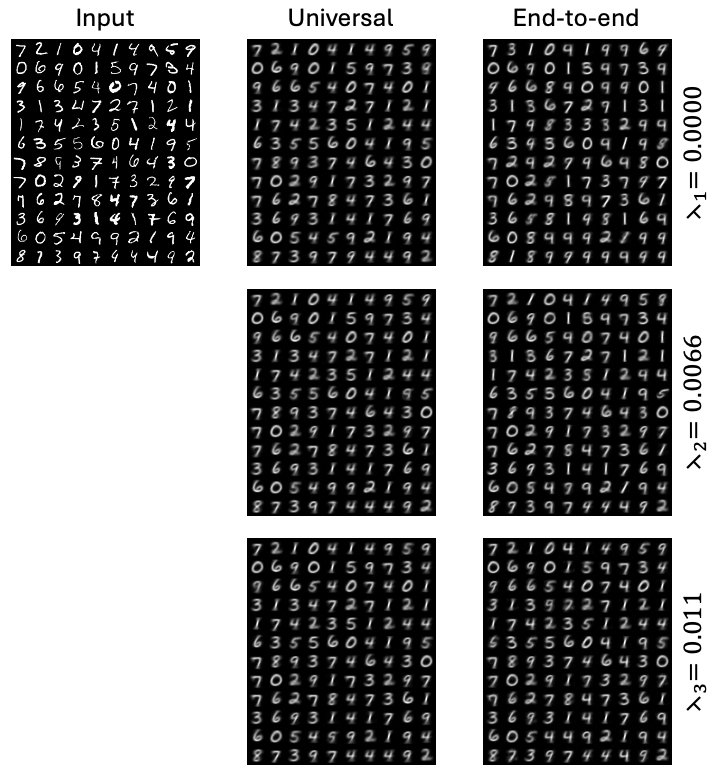}
    \caption{Decompression outputs of selected models for classification-distortion-rate at rate $R = 4.75$ on MNIST dataset. As the emphasis on classification loss ($\lambda_c$ increases), the decompression images become sharper.}
    \label{fig:CDR_Comparision_MNIST_1}
\end{figure}

\myheading{Setup and Training.} We employ a stochastic autoencoder architecture incorporating GAN and pre-trained classifier regularization. Specifically, the model comprises an encoder ($f$), decoder ($g$), critic ($h$), and a pre-trained classifier ($c$). Detailed network configurations are provided in the supplementary material.

Given an input image $x$, the encoder's final layer applies a tanh activation, producing a continuous representation $f(x) \in [-1,1]^d$. This representation is quantized into $L$ evenly spaced length intervals $2/(L-1)$ across each dimension. Employing the shared randomness assumption, we utilize dithered quantization \cite{gray1993dithered, theis2021advantages}, where both sender and receiver share a common random vector $u \sim U[-1/(L-1),+1/(L-1)]^d$. The sender computes:
\begin{equation}
    z = \text{Quantize}(f(x)+u),
\end{equation}
and transmits $z$. The receiver reconstructs the input by decoding $z-u$. During training, the soft gradient estimator from \cite{mentzer2018conditional} is used to propagate gradients through the quantization step.

Our training approach closely follows \cite{blau2019rethinking}. Initially, \textit{end-to-end} models are trained where the encoder, decoder, and critic are jointly optimized. We use mean squared error (MSE) as the distortion metric, Wasserstein-1 distance for perceptual realism, and classification accuracy derived from the pre-trained classifier $c$. Specifically, the reconstructed image $\hat{x}$ is classified by $c$, yielding predicted class probabilities $\hat{\mathbf{s}}$. Classification accuracy is determined using the true label $s$ and $\hat{\mathbf{s}}$.

Conditional entropy constraints are approximated via the cross-entropy (CE) loss with a pre-trained classifier parameterized by $\psi$, following \cite{boudiaf2021unifying_cross_entropy, Wang2024}:
\begin{align*}
    H(S|\hat{X}) &\leq \text{CE}(s,\mathbf{\hat{s}}).
\end{align*}
The compression rate $R$ is set as $\text{dim}\times \log_2(L)$, the product of the representation dimension and quantization levels, simplifying the implementation with minimal suboptimality \cite{blau2019rethinking, agustsson2019generative}.

The overall loss functions for the RDC and RDP cases are:
\begin{align}
\mathcal{L}_{RDC} &= \mathbb{E}[\|X - \hat{X}\|^2] + \lambda_c \text{CE}(s, \mathbf{\hat{s}}), \\
\mathcal{L}_{RDP} &= \mathbb{E}[\|X - \hat{X}\|^2] + \lambda_p W_1(p_X, p_{\hat{X}}),
\end{align}
where $\lambda_c$ and $\lambda_p$ control the tradeoffs between distortion, classification, and perception.

To build universal models, we freeze encoders from previously trained end-to-end models and retrain only new decoders ($g_1$) and critics ($h_1$) using modified parameters ($\lambda^1_c$, $\lambda^1_p$):
\begin{align}
\mathcal{L}^1_{RDC} &= \mathbb{E}[\|X - \hat{X}_1\|^2] + \lambda^1_c \text{CE}(s, \mathbf{\hat{s}}_1), \\
\mathcal{L}^1_{RDP} &= \mathbb{E}[\|X - \hat{X}_1\|^2] + \lambda^1_p W_1(p_X, p_{\hat{X}_1}).
\end{align}
Weights of the new decoders and critics are initialized randomly or from the previously trained critic. This process is repeated over multiple parameters ($\lambda^i_c$, $\lambda^i_p$) to produce comprehensive RDC and RDP tradeoff curves. Additional details are included in the supplementary material.

\subsection{Results}
Figure \ref{fig:classification_rate_distortion} presents the classification-distortion-rate tradeoffs for MNIST using several rates defined by $R=\text{dim}\times\log_2(L)$ with dimension-level pairs $(3,3)$, $(3,4)$, and $(4,4)$. Each point corresponds to a trained encoder-decoder pair at specific combinations of $R$ and $\lambda_c$. Points sharing the same color indicate identical rates, clearly illustrating the inherent tradeoff: enhancing classification accuracy generally leads to increased distortion. Additionally, higher rates enable simultaneous improvements in distortion and classification accuracy.

Figure \ref{fig:PDR_Comparision_MNIST_1} demonstrates the performance comparison between end-to-end and universal models on the rate-distortion-perception curve at a fixed rate of $R = 4.75$. Universal models, which reuse an encoder trained from an end-to-end model, closely approach the performance of their end-to-end counterparts across the perception-distortion tradeoff, validating prior observations \cite{UniversalRDPs}.

Similarly, Figure \ref{fig:CDR_Comparision_MNIST_1} presents the rate-distortion-classification curve for the same fixed rate ($R = 4.75$). Notably, we observe a significant distortion gap when reusing encoders trained at a particular distortion-classification tradeoff for other tradeoff points. To address this, we recommend selecting the tradeoff parameter $\lambda^1_c$ at a scale different from the original $\lambda_c$, allowing universal models to effectively span the entire distortion-classification tradeoff spectrum.

\section{Conclusion}
In this work, we examined the inherent tradeoffs among rate, distortion, classification accuracy, and perceptual quality within lossy compression frameworks. While previous approaches typically involved training end-to-end systems separately for each desired tradeoff point, our findings suggest that fixing a carefully optimized representation (encoder) and only adapting the decoder is sufficient to achieve comparable performance. This significantly simplifies system design by reducing computational costs and model redundancy. Future research may explore extending this universal representation concept to more complex scenarios, including high-resolution image and video compression tasks.

{\small
\bibliographystyle{ieee_fullname}
\bibliography{egbib}
}    

\appendix

\section{Architecture}
The architecture employed in our experiments is a stochastic autoencoder combined with GAN and pre-trained classifier regularization. Each model is composed of an encoder, decoder, critic, and classifier. The detailed network architectures for these components are summarized in Table \ref{tab:architecture}, with each row indicating a sequence of layers.

\begin{table}[h]
    \centering
    \caption{Detailed architecture for encoder, decoder, critic, and pre-trained classifier used in MNIST experiments. l-ReLU denotes Leaky ReLU activation.}
    \label{tab:architecture}
    \begin{tabular}[t]{|c|}
    \multicolumn{1}{c}{Encoder} \\
	\hline 
        Input  \\ 
        \hline 
        Flatten  \\
        \hline 
        Linear, BatchNorm2D, l-ReLU \\
        \hline 
        Linear, BatchNorm2D, l-ReLU \\
        \hline 
        Linear, BatchNorm2D, l-ReLU \\
        \hline 
        Linear, BatchNorm2D, l-ReLU \\
        \hline 
        Linear, BatchNorm2D, Tanh \\
        \hline
        Quantizer \\
    \hline
    \end{tabular}
    \quad
    \begin{tabular}[t]{ |c| }
    \multicolumn{1}{c}{Decoder} \\
	\hline 
        Input  \\
        \hline 
        Linear, BatchNorm1D, l-ReLU \\
        \hline 
        Linear, BatchNorm1D, l-ReLU \\
        \hline
        Unflatten \\ 
        \hline 
        ConvT2D, BatchNorm2D, l-ReLU \\
        \hline 
        ConvT2D, BatchNorm2D, l-ReLU \\
        \hline
        ConvT2D, BatchNorm2D, Sigmoid \\
    \hline
    \end{tabular}
    \quad
    \begin{tabular}[t]{ |c| }
    \multicolumn{1}{c}{Critic} \\
	\hline 
        Input  \\ 
        \hline 
        Conv2D, l-ReLU \\
        \hline 
        Conv2D, l-ReLU \\
        \hline
        Conv2D, l-ReLU \\
        \hline
        Linear  \\  
    \hline
    \end{tabular}
    \quad
    \begin{tabular}[t]{ |c| }
    \multicolumn{1}{c}{Pre-trained Classifier} \\
	\hline 
        Input  \\
        \hline 
        Conv2D (10 filters, kernel=5), ReLU \\
        \hline 
        MaxPool2D (kernel size=2) \\
        \hline
        Conv2D (10 filters, kernel=5), ReLU \\
        \hline
        MaxPool2D (kernel size=2) \\
        \hline
        Flatten \\
        \hline
        Linear, ReLU \\
        \hline
        Linear, Softmax \\
    \hline
    \end{tabular}
\end{table}

Table \ref{tab:hyperparameters} summarizes the hyperparameters used in training each component of the model across all experiments.

\begin{table}[!htb]
    \centering
    \caption{Training hyperparameters used across all experiments.}
    \label{tab:hyperparameters}
    \begin{tabular}{ccccc}
    \specialrule{.1em}{.05em}{.05em} 
    & $\alpha$ & $\beta_1$ & $\beta_2$ & $\lambda_{\text{GP}}$ \\ 
    \hline
    Encoder & $10^{-2}$ & $0.5$ & $0.9$ & - \\
    Decoder & $10^{-2}$ & $0.5$ & $0.9$ & - \\
    Critic & $2 \times 10^{-4}$ & $0.5$ & $0.9$ & $10$ \\
    Classifier & $10^{-3}$ & $0.9$ & $0.999$ & - \\
    \hline
    \end{tabular}
\end{table}

\end{document}